\documentclass[runningheads]{llncs}
\usepackage[T1]{fontenc}
\usepackage{graphicx}
\usepackage{multirow}%
\usepackage{amsmath,amssymb,amsfonts}%
\usepackage{mathrsfs}%
\usepackage{tabularx,booktabs}
\usepackage[title]{appendix}%
\usepackage{xcolor}%
\usepackage{textcomp}%
\usepackage{manyfoot}%
\usepackage{booktabs}%
\usepackage{algorithm}%
\usepackage{algorithmicx}%
\usepackage{algpseudocode}%
\usepackage{listings}%
\usepackage{orcidlink}
\begin{document}
\title{Bengali Text Classification: An Evaluation of Large Language Model Approaches}
\titlerunning{Bengali Text Classification: An Evaluation of LLM Approaches}
%
\author{Md Mahmudul Hoque$^*$\inst{1}\orcidlink{0000-0002-2618-4157} \and
Md Mehedi Hassain\inst{2}\orcidlink{{0009-0000-0818-245X}} \and
Md Hojaifa Tanvir\inst{3}
\and
Rahul Nandy \inst{2}}
\authorrunning{M. M. Hoque et al.}
%
\institute{Dept. of Computer Science \& Engineering, CCN University of Science and Technology, CCN Road, Cumilla - 3506, Bangladesh. \\ 
\email{cse.mahmud.evan@gmail.com$^*$}\\
\and
Dept. of Electrical \& Electronic Engineering, International Islamic University Chittagong,Chattogram - 4318, Bangladesh. \\
\email{\{mdmehedihasan225588,rnandy107\}@gmail.com}\\
\and
Dept. of Computer Science \& Engineering,University of Asia Pacific, Dhaka - 1209, Bangladesh.\\
\email{hojaif.tanvir@gmail.com}}
\maketitle              
\begin{abstract}
Bengali text classification is a Significant task in natural language processing (NLP), where text is categorized into predefined labels. Unlike English, Bengali faces challenges due to the lack of extensive annotated datasets and pre-trained language models. This study explores the effectiveness of large language models (LLMs) in classifying Bengali newspaper articles. The dataset used, obtained from Kaggle, consists of articles from Prothom Alo, a major Bangladeshi newspaper. Three instruction-tuned LLMs—LLaMA 3.1 8B Instruct, LLaMA 3.2 3B Instruct, and Qwen 2.5 7B Instruct—were evaluated for this task under the same classification framework. Among the evaluated models, Qwen 2.5 achieved the highest classification accuracy of 72\%, showing particular strength in the "Sports" category. In comparison, LLaMA 3.1 and LLaMA 3.2 attained accuracies of 53\% and 56\%, respectively. The findings highlight the effectiveness of LLMs in Bengali text classification, despite the scarcity of resources for Bengali NLP. Future research will focus on exploring additional models, addressing class imbalance issues, and refining fine-tuning approaches to improve classification performance.

\end{abstract}
\section{Introduction}
Text classification \cite{exxactcorpWhatText} refers to the process of assigning predefined categories or labels to text based on its content. It is a supervised machine learning task where an algorithm is trained on a labeled dataset to recognize patterns and features in the text that correspond to particular categories. Once trained, the model can classify unseen text. In natural language processing \cite{ria2020toward}, text classification is a key area of research. Using Natural Language Processing (NLP), text classifiers first analyze a given text and then categorize it according to the dataset. Text classification, \cite{striveworksTextClassification} used widely for spam detection, sentiment analysis, and other purposes, is one of the most fundamental applications of NLP. The strength of LLM in NLP may enable them to perform the task at a high level. The flexibility \cite{mediumWhenWould} of the pre-trained LLM method is highly beneficial for tasks that require ad hoc or experimental classification. It is particularly useful in scenarios where the task is performed infrequently or for a single activity, the classes may change over time, or when more detailed output beyond just a class label is desired.

Our research aims to perform text classification on a comprehensive Bengali newspaper dataset comprising over 400,000 samples and 25+ categories in this newspaper data.The research focuses on classifying Bengali news content into 9 significant categories. The LLaMA 3.1-8B-Instruct model, LLaMA 3.2-3B-Instruct model, and Qwen 2.5 7B-Instruct model were utilized to develop a comprehensive framework capable of accurately classifying Bengali text. To ensure the accuracy of the findings, the model’s performance was evaluated using metrics such as accuracy, F1 score, recall, precision, and confusion matrices. The research also aims to compare three large language models for Bengali newspaper text classification, providing insights and identifying potential research directions in this domain. The contributes to the development of Bengali text classification models, specifically utilizing Large Language Models (LLMs). By focusing on key categories within Bengali news articles, the research aims to establish a framework that can serve as a foundation for future researchers to develop more efficient and accurate text classification systems for Bengali.
The Bengali language, spoken by more than 267 million people \cite{lingostarBengaliLanguage}, is significantly underrepresented in the field of Natural Language Processing (NLP). Despite its widespread use, there is a notable lack of resources and research dedicated to Bengali text classification. Recent advancements in Large Language Models (LLMs) have demonstrated their potential in various NLP tasks, yet their application to Bengali remains largely unexplored.  As Bengali remains a relatively underexplored language in the field of natural language processing, this study paves the way for further innovations.

\section{Related Work}
The authors in \cite{vulpescu2024optimized} developed a fine-tuned LLaMA model that performs well even with limited data and results showed over 87\% accuracy in topic categorization and a significant reduction in hallucinations, where models produce incorrect or irrelevant outputs. Analyzing large language models (LLMs) for classifying Spanish legislative texts \cite{pena2023leveraging}, they used a refined dataset of 33,000 documents annotated for 30 topics and demonstrated the potential of LLMs in handling complex, domain-specific tasks. Future work aims to address dataset biases and explore advanced NLP and multimodal methods for legislative analysis. By examining topic detection in news texts \cite{kosar2024comparative} through two experiments—one involving 109 participants across three countries and another comparing eight native speakers with six LLMs—the study demonstrated valuable insights. The findings revealed substantial variation in human topic categorization, while advanced LLMs, particularly GPT-4, performed equally well or better. The study emphasized the need for improved evaluation methods and more diverse datasets in future research. TELEClass, a novel minimally-supervised hierarchical text classification method, was proposed in \cite{zhang2024teleclass}. It introduces two key innovations: improving label taxonomy using class-indicative terms derived from LLMs and the corpus, and utilizing LLMs for data annotation and generation. Evaluated on two public datasets, TELEClass outperformed baseline models, as confirmed by ablation studies. The study also examined the effectiveness of zero-shot LLM prompting for hierarchical classification. Future research will focus on adapting TELEClass to low-resource text mining tasks and expanding its ability to handle more complex label hierarchies.

In \cite{el2024comparative}, the authors evaluation of the interpretability of three topic modeling approaches— Latent Dirichlet Allocation (LDA), BERTopic, and RoBERTa—across three case studies demonstrated that LLM-based methods (BERTopic and RoBERTa) consistently achieved superior interpretability compared to the traditional LDA. This enhanced performance was attributed to the LLM-based models' ability to generate more coherent topics while mitigating issues like feature sparsity and unclear semantic information. While the results favored modern LLM approaches, the authors suggested future research could expand the comparison to include other probabilistic models such as LSA and PLSA. Moreover, A classification model development approach \cite{ria2020toward} was proposed using several machine learning algorithms, including Naïve Bayes (NB), XGBoost (XGB), Random Forest (RF), K-Nearest Neighbors (KNN), Support Vector Classifier (SVC), and Decision Tree (DT). Using a dataset of over thousand sentences in standard and colloquial forms, their study found Naïve Bayes achieved optimal accuracy at 77\%. The research identified computational limitations and dataset size constraints as key challenges. The authors suggested that expanding the dataset and enhancing computational resources could improve model performance, noting that larger datasets typically yield better predictive accuracy in natural language processing tasks. Another systematic approach to automatic text classification was proposed in \cite{liang2024research}, focusing on the feature extraction and classification processing stages. The research emphasized the critical role of feature selection in identifying relevant textual attributes and explored various machine learning algorithms for effective classification. The study demonstrated that carefully selected feature extraction techniques and classification models significantly impact system performance. The authors recommended future research directions, including the integration of advanced approaches like deep learning and transformer-based models to handle increasingly complex text data and improve classification accuracy. A novel text classification approach for\cite{dien2019article} article categorization was developed, incorporating feature extraction, data vectorization, and multiple machine learning algorithms. The method was evaluated using Support Vector Machines (SVM), Naïve Bayes, and k-Nearest Neighbors on two distinct article datasets. Results demonstrated superior performance, with SVM achieving over 91\% accuracy. While the findings validated the effectiveness of combining natural language processing with machine learning for automatic article classification, the authors acknowledged dataset size limitations and recommended future research with larger, more diverse datasets and advanced processing techniques. Another innovative text classification model was introduced in \cite{hossain2021bengali}, combining GloVe embeddings with Very Deep Convolutional Neural Networks (VDCNN) for Bengali text classification. Two essential resources were created: an Embedding Corpus of 969,000 unlabeled texts and a Bengali Text Classification Corpus with 156,207 labeled documents. Their novel Embedding Parameters Identification Algorithm evaluated 165 embedding models, determining GloVe's superiority over Word2Vec, FastText, and m-BERT. The proposed GloVe + VDCNN model achieved 96.96\% accuracy on the classification corpus, demonstrating significant advancement in Bengali text classification and establishing a framework for low-resource language processing.

Furthermore, The under-resourced \cite{rezaul2020classification} status of Bengali in NLP was addressed by introducing three datasets for hate speech detection, document classification, and sentiment analysis. BengFastText, the largest Bengali word embedding model, was developed, trained on 250 million articles. Using a Multichannel Convolutional LSTM (MConv-LSTM) network, they achieved notable F1-scores: 92.30\% for document classification, 82.25\% for sentiment analysis, and 90.45\% for hate speech detection. The study demonstrated BengFastText's superior performance over Word2Vec and GloVe, highlighting its capability to capture semantic nuances. This work provides crucial benchmarks and resources, advancing Bengali NLP and encouraging further exploration with larger datasets and complex models. Another advancements \cite{roy2023bengali} in Bengali text classification were made by applying both machine learning and deep learning techniques.Their primary objective was to bridge the gap between English and Bengali text classification, as the majority of research in text classification has historically focused on English, leaving Bengali significantly underexplored. While text classification in English has gained substantial attention in the research community, there remains a limited amount of work addressing the unique challenges and opportunities of Bengali text classification. To address this issue, they introduced a new and comprehensive dataset specifically designed for Bengali text classification. Clustering Techniques \cite{hoque2024comparison} used for comparing two well-known and respected English and Bengali daily newspapers in Bangladesh. Author introduced a revenue generation model for two most popular newspapers in Bangladesh. Developing resources and demonstrating \cite{sazzed2020cross} the effectiveness of cross-lingual sentiment analysis for Bengali focused on addressing the lack of sufficient resources for sentiment analysis in the Bengali language, leading to the creation of a comprehensive and annotated dataset. The authors developed a corpus consisting of approximately 12,000 Bengali reviews, specifically designed to serve as a benchmark for sentiment analysis tasks in Bengali and compared the performance of ML classifiers trained on the original Bengali corpus and the translated English version, using metrics like Cohen’s Kappa to assess sentiment preservation. Results highlighted the potential of machine translation for cross-lingual sentiment analysis, offering a viable approach for under-resourced languages. This research provides valuable resources and insights, advancing Bengali sentiment analysis. Another bengali word embeddings \cite{ahmad2016bengali} were developed using the Word2Vec model and applied to news document classification.By analyzing Bengali news articles from the past five years, they created semantic word representations, which were clustered using the K-means algorithm to group similar words. For classification, a Support Vector Machine (SVM) achieved a high F1-score of 91\%, demonstrating the embeddings' effectiveness in capturing word semantics. This study highlights the potential of word embeddings to enhance Bengali NLP, addressing resource limitations and providing a foundation for applying similar methods to other text-based tasks in under-resourced languages. Transformer models for Bangla text classification were explored \cite{alam2020bangla}, demonstrating their superiority over traditional machine learning and deep learning approaches like CNNs and LSTMs. Fine-tuning pre-trained transformers achieved state-of-the-art results across six benchmark datasets, with accuracy improvements of 5-29\% depending on the task. These models excelled at capturing nuanced contextual relationships in Bangla text, highlighting their effectiveness for under-resourced languages. The study underscores the potential of transformers to advance Bangla NLP and encourages further adoption of these models for diverse applications in low-resource languages, contributing to the broader development of language processing in such contexts. Several machine learning algorithms—SVM, MNB, SGD, and LR—were evaluated in \cite{tudu2018performance} for Bengali text classification using a publicly available corpus of 84,906 Bengali newspaper articles across 9 categories. They studied the impact of training dataset size on model accuracy and observed how varying data volumes affected performance. Additionally, they recorded training times to assess the computational efficiency of each model. The study highlighted the unique challenges of Bengali text classification, due to limited language-specific resources, and provided valuable benchmarks and insights for future research in this area. Another convolutional neural network (CNN) model with FastText embeddings was proposed in \cite{hossain2020text} for text document classification in resource-constrained languages. The system, consisting of embedding generation, feature representation, training, and testing, was evaluated on datasets of 39,079 training and 6,000 validation texts, achieving 96.85\% accuracy on 9,779 test samples. Hyperparameter tuning improved classification performance. The study suggested expanding the model to include more document classes and exploring other embedding techniques like ElMo and BERT for future research, aiming to further enhance the system’s capabilities. Multiple time series models \cite{hoque2023analyzing} applied  to forecast monthly prices of grocery products across Bangladeshi divisions. Using MAE for evaluation, RNN and a Naïve Ensemble model showed the best performance in capturing price fluctuations. Deep learning approaches \cite{rahman2020bangla} have been effectively employed to classify Bangla text documents, utilizing both Convolutional Neural Networks (CNN) and Long Short-Term Memory (LSTM) networks for this classification task. Evaluating models on three datasets, the LSTM model achieved a high accuracy of 95.42\% on the Prothom Alo dataset. The study highlighted the advantages of character-level encoding over traditional document-level approaches in enhancing classification performance. Future work aims to develop hybrid models incorporating advanced techniques like Bi-directional LSTMs and BERT to further improve Bangla text classification, advancing deep learning in NLP for under-resourced languages.

\section{Methodology}
\begin{figure}[ht!]
    \centering
    \includegraphics[width=\linewidth]{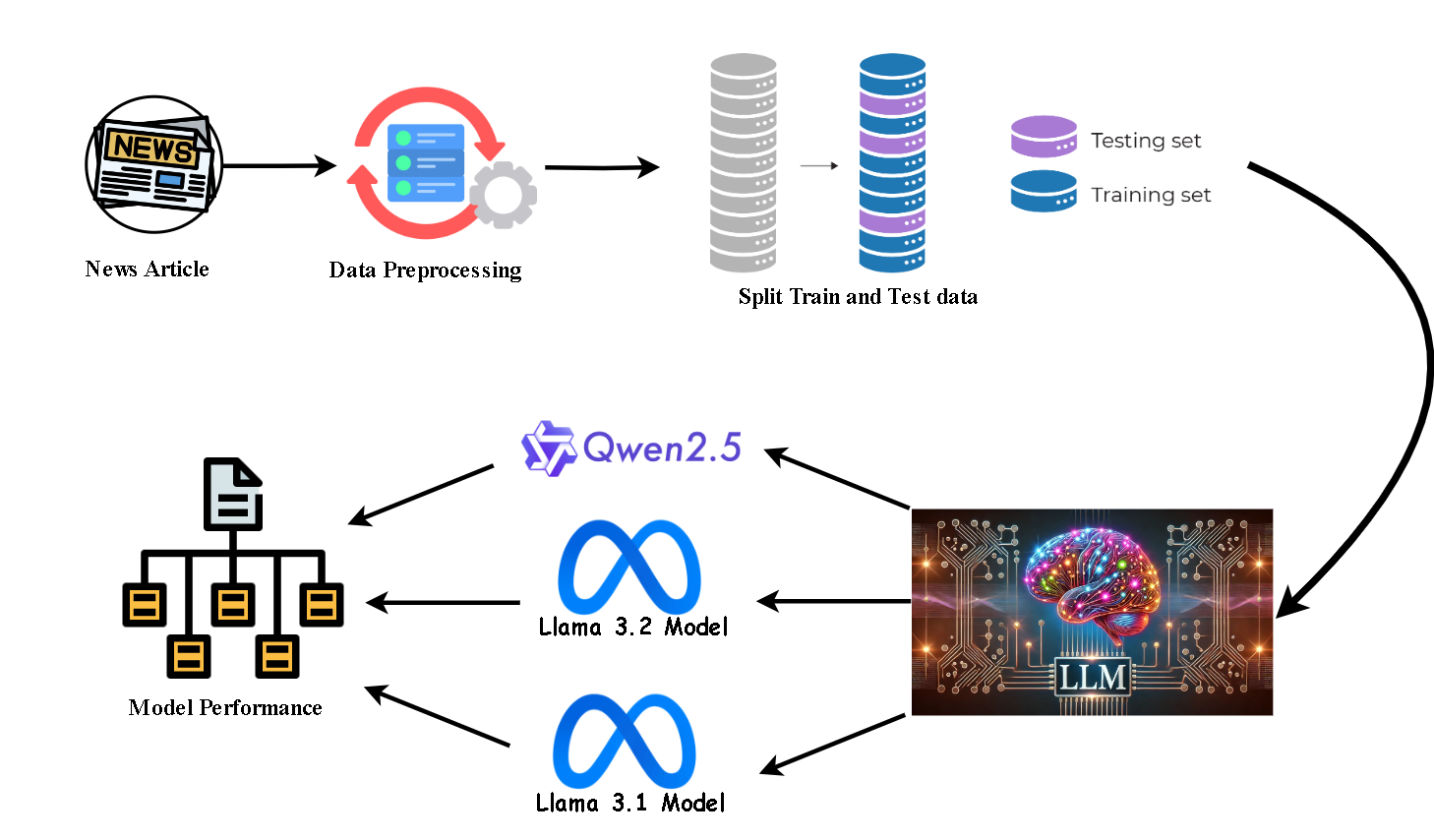}
    \caption{System Design}
    \label{system}
\end{figure}

This methodology presents a proposed system architecture of the text classification using Large Language Models and make a comparison between them. Fig \ref{system} shows the system architecture of the work.

\subsection{Data Collection}
\begin{figure}[ht!]
    \centering
    \includegraphics[scale=0.80]{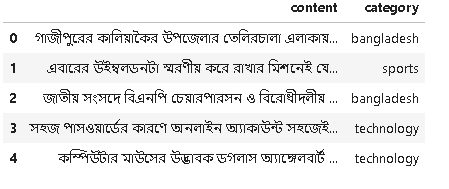}
    \caption{Data Sample}
    \label{data}
\end{figure}

The dataset \cite{zabir_al_nazi_2020} utilized in our research was sourced from Kaggle, an online platform renowned for its extensive collection of data sets and resources tailored for data science professionals and enthusiasts.
The specific dataset comprises a substantial corpus of Bangla newspaper articles from the Prothom Alo archive, one of Bangladesh’s leading newspapers \cite{prothomaloAbout}. This data set includes over 400,000 news samples, classified into more than 25 distinct categories, offering a wide variety of content. The research utilized the second version of the dataset, which consists of 9 attributes and a total of 437,948 samples. However, in this study we used only 2 attributes to classify the text. The data of the dataset is illustrated in Fig \ref{data}.

\subsection{Data Preprocessing}
 
\begin{figure}
    \centering
    \includegraphics[width=.80\linewidth,trim =20 32 50 20,clip]{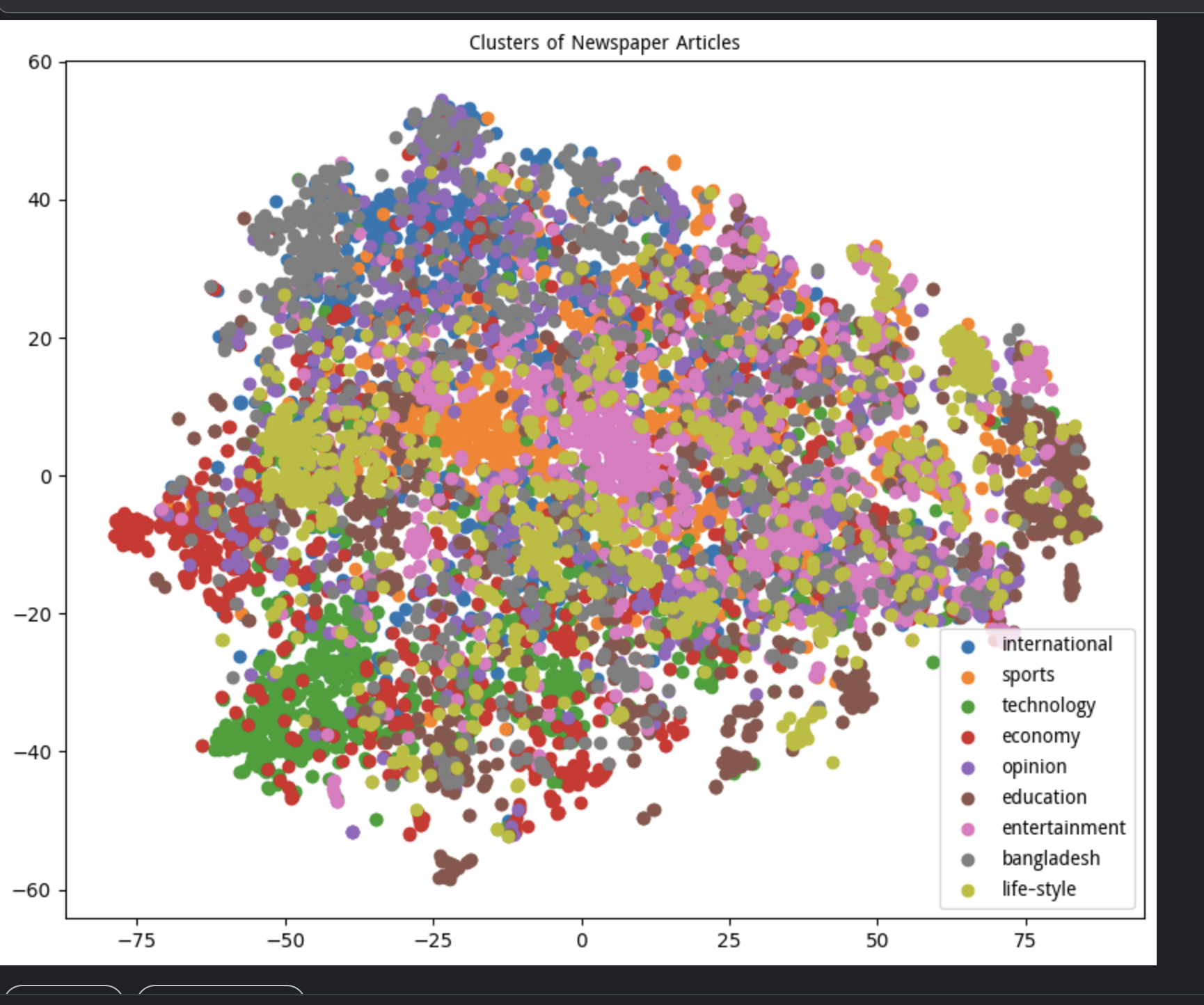}
    \caption{Clusters of neighbors News Articles}
    \label{cls}
\end{figure}

The dataset consists of 9 attributes and 437,948 samples. During the analysis, duplicate samples were identified and removed to ensure data integrity. For the text and topic classification task, the focus was on two specific columns: ‘Content’ and ‘Category’. However, an issue of class imbalance was observed in the ‘Category’ column. Class imbalance poses a significant challenge in classification tasks, as it can lead to biased models that favor the majority class. To address this, the Random Under Sampler technique was employed. This method balances the data set by randomly deleting rows from the majority class(es) based on the sampling strategy. Random Under Sampler provides a quick and efficient way to achieve a balanced data set by selecting a representative subset of data for the targeted classes \cite{mediumUnderSamplingMethods}. In addition, a clustering approach using KNN was performed to identify the nearest neighbors of the articles and their corresponding categories. Figure \ref{cls} illustrates the clustering neighbors of the news articles and their categories.

\subsection{Model Development}
In our research three Large Language Models (LLM) to classify Bengali news text which are LLaMA (3.1, 3.2) and Qwen 2.5.  We employed the LLaMA 3.1 and 3.2 models developed by Meta, both recognized for their enhanced performance and scalability. For LLaMA 3.1, we used the '8B Instruct' variant, which includes 8 billion parameters, while for LLaMA 3.2, we leveraged the '3B Instruct' variant with 3 billion parameters. These models offer a balance between computational efficiency and predictive accuracy. Additionally, we integrated Transformer version 2 to optimize text classification outcomes\cite{llama,huface}. We compare LLaMA models with Qwen 2.5 7B Instruct model, which is developed by Alibaba Cloud. It based on transformer with RoPE, SwiGLU, RMSNorm, and Attention QKV bias. Having 28 layers, it has 7.61 billion parameters \cite{QwenQwen25}. The implementation utilized a 4-bit quantization type and a float16 computation type. The llm\_int8\_enable\_fp32\_cpu\_offload feature was enabled to mitigate overflow issues, facilitating the accommodation of large models and enabling distribution between CPU and GPU. The device map was set to ‘auto’ to automatically allocate the model’s layers across available devices. AutoTokenizer was employed to convert texts into tokens, a process that involves splitting the text into smaller units (tokens), which can be words, subwords, or characters, depending on the tokenizer’s configuration. The model architecture involved training a language model using LoRA (Low-Rank Adaptation) and QLoRA (Quantized Low-Rank Adaptation) techniques. This approach significantly reduced the number of trainable parameters. The training process utilized a cosine learning rate scheduler and the AdamW optimizer. We employed gradient accumulation and checkpointing techniques.Metrics were reported to Weights \& Biases (wandb) for comprehensive tracking. To ensure precise text classification, we developed a well-structured prompt that directs the model to classify news articles into one of nine specific categories: Bangladesh, Sports, Technology, Entertainment, International, Economy, Life-Style, Opinion, and Education. This prompt was consistently applied throughout the model training, helping the LLM effectively associate the content with the appropriate category and enhancing the model’s ability to classify text accurately. 
\\
\textbf{Device Requirements:}\\
CPU: Core i3 5th Gen\\
GPU: GPU P100\\
RAM: 32GB\\
Run Time: 3751.8 s , 3495.4 s and 8105.6 s

\section{Results and Discussions} 
The performance of the fine-tuned model was evaluated for classifying Bengali newspaper articles across multiple categories. Key metrics such as accuracy, precision, recall, F1-score, and a confusion matrix were analyzed. Additionally, a comparison between two popular LLMs—LLaMA and Qwen—was conducted. The dataset was split into 80\% for training and 20\% for testing and validation. For visualization, Plotly Express and Seaborn were used to generate insightful graphs and charts. Qwen 2.5 7B-Instruct  performed the best with an accuracy of 72\%, demonstrating its effectiveness in capturing semantic patterns in Bengali text. This model's enhanced performance is likely attributed to its larger number of parameters and advanced architectural features, which enable it to effectively process the complexities of Bengali language classification. In contrast, the LLaMA 3.1 8B-Instruct model exhibited a moderate accuracy of 53\%, while the LLaMA 3.2 3B-Instruct model achieved a slightly higher accuracy of 56\%. Although these models are smaller in terms of parameters, their performance remains reasonable but falls short of Qwen 2.5.  Table \ref{acc} shows the model accuracy and Table \ref{qwen} shows the model performance for each category news.

\begin{table}[htbp]
\caption{Model Accuracy}
\centering
\begin{tabular}{|p{2cm} | p{3cm}|} 
\hline
 \textbf{LLM} & \textbf{Accuracy(\%)}  \\ 
\hline
LLaMA 3.1 & 53\%  \\ \hline
LLaMA 3.2 & 56\%  \\ \hline
Qwen 2.5 & 72\%  \\ \hline

\end{tabular}
\label{acc}
\end{table}

\begin{table}[htbp]
\caption{Evaluation Metrics for Fine-Tuned Qwen Model }
\centering
\begin{tabular}{c c c c c }
\hline
\toprule
 \textbf{Category} & \textbf{Accuracy(\%)} & \textbf{Precision(\%)} & \textbf{Recall(\%)} & \textbf{F1-Score(\%) } \\ 
\midrule
Bangladesh & 73.0 & 70.0 & 74.0 & 72.0 \\ \hline
Sports & 81.0 & 84.0 & 81.0 & 83.0 \\ \hline
Technology & 76.0 & 79.0 & 76.0 & 78.0 \\ \hline
Entertainment & 75.0 & 74.0 & 75.0 & 75.0 \\ \hline
International & 74.0 & 67.0 & 74.0 & 70.0 \\ \hline
Economy & 69.4 & 76.0 & 79.0 & 73.0 \\ \hline
Life Style & 67.2  & 66.0 & 67.0 & 66.0 \\ \hline
Opinion & 58.8 & 60.0 & 59.0 & 59.0 \\ \hline
Education & 78.3 & 81.0 & 78.0 & 80.0 \\ \hline

\bottomrule
\end{tabular}
\label{qwen}
\end{table}

\begin{figure}[ht!]
    \centering
    \includegraphics[width=.88\linewidth]{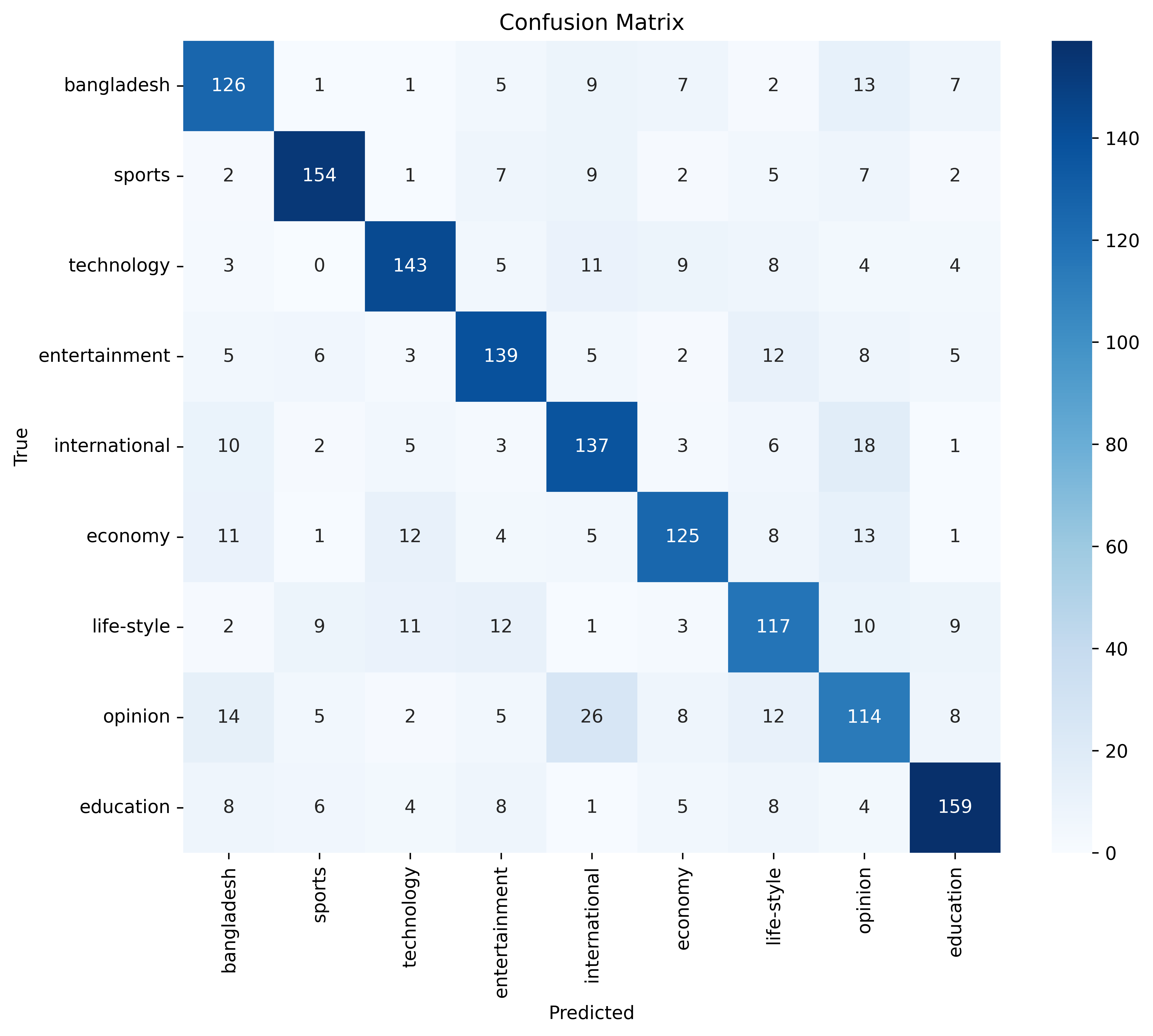}
    \caption{Confusion Matrix for Qwen Model}
    \label{122}
\end{figure}

In Fig \ref{122} demonstrates Confusion Matrix for fine tuned Qwen Model. Table \ref{qwen} presents the model evaluation metrics for each category of news. The model demonstrated strong performance across all categories, with 'Sports' achieving the highest accuracy, followed by 'Technology' and 'Entertainment'. Precision, recall, and F1-scores were also robust, reflecting the model's effectiveness in accurately classifying diverse content. While the 'Sports' category performed best, and got the slightly lower accuracy in 'Technology'. 

\subsection{Discussion}
The results highlight Qwen 2.5’s strong performance in Bengali news classification, while also underscoring key challenges in low-resource language processing. Qwen 2.5 7B-Instruct achieved the highest overall accuracy at 72\%, significantly outperforming LLaMA 3.1 8B-Instruct (53\%) and LLaMA 3.2 3B-Instruct (56\%). It excelled in categories like Sports (81\%), Technology (76\%), and Education (78.3\%), where consistent patterns and terminology likely aided classification. Sports emerged as the easiest to classify across all models, while Opinion, Life Style, and Economy posed greater difficulty due to their more subjective or varied language.
Qwen also showed balanced precision and recall, with F1-scores ranging from 59\% (Opinion) to 83\% (Sports), suggesting strong generalization across topics. Notably, the smaller LLaMA 3.2 slightly outperformed its larger counterpart, hinting that architecture may matter more than size. Quantization may have further impacted accuracy by introducing information loss.
Confusion Matrix revealed most misclassifications occurred between similar categories—like Economy and International, or Life Style and Entertainment—showing the models recognized semantic overlap rather than making random mistakes. Overall, while Qwen 2.5 demonstrates clear potential, the study also reveals areas for improvement in classifying nuanced or overlapping content. These findings offer a solid foundation for advancing Bengali text classification using LLMs.

\section{Conclusion}
This study focuses on fine-tuning Bengali news text classification using two prominent large language models (LLMs), LLaMA and Qwen, and compares their performance. The dataset sourced from Kaggle, comprising news articles from Prothom Alo, a leading Bangladeshi newspaper, was utilized for this research. The dataset contains over 400,000 samples across more than 25 categories, but for the purpose of this study, we focused on nine of the most popular categories to streamline the classification task. The initial phase involved the collection and preprocessing of the data, which was essential for cleaning and structuring the dataset to enhance model performance. During preprocessing, irrelevant attributes were removed, and the data was reformatted to be suitable for model training. Subsequently, we fine-tuned the LLaMA (3.1 and 3.2) and Qwen 2.5 models, all of which are optimized for text classification tasks with 8B, 3B, and 7B parameters, respectively. These models, with their advanced transformer architecture, were applied to the Bengali dataset. Following fine-tuning, the LLaMA 3.1 model achieved an accuracy of 53\%, the LLaMA 3.2 model attained 56\% accuracy, and the Qwen 2.5 model outperformed the others with a 72\% accuracy rate. Although these results indicate the models' capability to classify Bengali text, there is scope for further enhancement through additional fine-tuning and optimization techniques. The observed class imbalance, with one category having significantly more data, may have impacted model's performance. Future work will focus on addressing this imbalance and optimizing the models for more accurate classification across all categories. Additionally, the dataset will be expanded by incorporating additional sources to improve its diversity and comprehensiveness. Future research will also explore a broader range of LLMs to identify models that can further enhance classification accuracy and overall performance.

%
%
%
\bibliographystyle{splncs04}
\bibliography{mybibliography}
%




\end{document}